\documentclass[10pt,twocolumn,letterpaper]{article}

\usepackage{iccv}
\usepackage{authblk}
\usepackage{times}
\usepackage{epsfig}
\usepackage{graphicx}
\usepackage{amsmath}
\usepackage{amssymb}
\usepackage{enumitem}
\usepackage{multirow}
\usepackage{booktabs}
\usepackage{bbm}
\usepackage[T1]{fontenc}
\usepackage[utf8]{inputenc}
\usepackage[english]{babel}
\usepackage[autostyle, english=american]{csquotes}
\usepackage{subcaption}
\usepackage{ulem}

\MakeOuterQuote{"}
\usepackage{bbm}

\usepackage[breaklinks=true,bookmarks=false]{hyperref}

\iccvfinalcopy 

\def\iccvPaperID{****} 

\ificcvfinal\fi
\ificcvfinal\fi
\begin{document}

\title{Solution for Point Tracking Task of ICCV 1st Perception Test Challenge 2023}


\author{

Hongpeng Pan\textsuperscript{1},
Yang Yang\textsuperscript{1,$\thanks{Corresponding Author}$} ,
Zhongtian Fu\textsuperscript{1},
Yuxuan Zhang\textsuperscript{1},
Shian Du\textsuperscript{2},
Yi Xu\textsuperscript{3},
Xiangyang Ji\textsuperscript{2}
}

\affil{

 $^1$Nanjing University of Science and Technology
 $^2$Tsinghua University
 $^3$Dalian University of Technology
}
\maketitle
\setlength{\intextsep}{1pt}
\setlength{\abovecaptionskip}{1.5pt}

\begin{abstract}

 This report proposes an improved method for the Tracking Any Point (TAP) task, which tracks any physical surface through a video. Several existing approaches have explored the TAP by considering the temporal relationships to obtain smooth point motion trajectories, however, they still suffer from the cumulative error caused by temporal prediction. To address this issue, we propose a simple yet effective approach called TAP with confident static points (TAPIR+), which focuses on rectifying the tracking of the static point in the videos shot by a static camera. To clarify, our approach contains two key components: (1) Multi-granularity Camera Motion Detection, which could identify the video sequence by the static camera shot. (2) CMR-based point trajectory prediction with one moving object segmentation approach to isolate the static point from the moving object. Our approach ranked first in the final test with a score of 0.46.

\end{abstract}

\vspace{-5pt}
\section{Introduction}

\begin{figure}[ht]
    \centering
    \begin{subfigure}[t]{0.45\linewidth}
        \includegraphics[width=\textwidth]{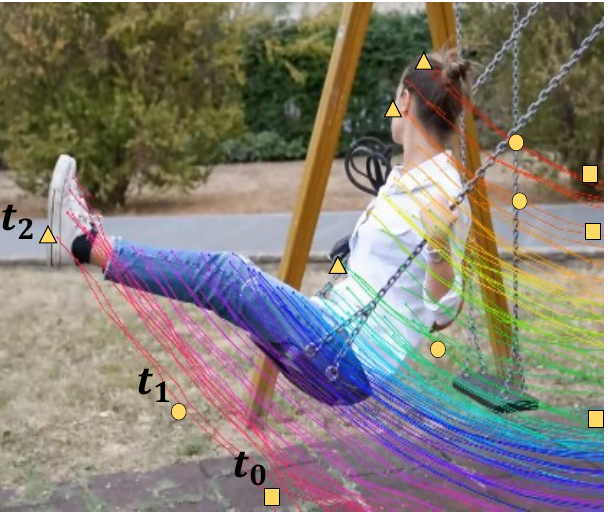}
        \parbox{\textwidth}{\centering ({\it a}) {Points Tracking Task}}     
    \end{subfigure}
    \begin{subfigure}[t]{0.45\linewidth}
        \includegraphics[width=\textwidth]{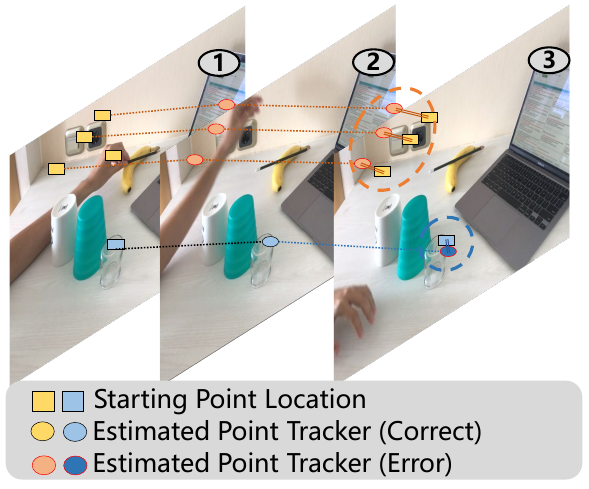}
        \parbox{\textwidth}{\centering ({\it b}) {Problems in Task}}     
    \end{subfigure}
    \caption{In (a), points of different shapes represent prediction results at different moments ($t_0$, $t_1$, $t_2$). At label 3 in (b), the blue dashed box illustrates the phenomenon of jitter around the stationary point, while the orange dashed box reflects the issue of static point pseudo-following. The two phenomena of positional offset occurring when the model predicts these static points are significant factors limiting the model's performance.}
    \label{fig: problem}
\end{figure}

Deep learning techniques have attracted widespread attention in multiple research fields~\cite{0074ZGGZ22,YangHGXX23,YangWZX019,YangZZXJY23}. Integrated into single-point tracking, which focuses on locating and following a specific object or feature across consecutive frames, they play a critical role in computer vision for tasks such as scene understanding and action recognition. As shown in Figure \ref{fig: problem} (a), it entails identifying whether two pixels from different images of the same video correspond to the same point on a physical surface. In this competition, we adopted a zero-shot strategy and explored three approaches: OmniMotion\cite{abs-2306-05422}, TAPIR\cite{abs-2306-08637}, and Cotraker\cite{abs-2307-07635}. The experimental results show that OmniMotion and Cotraker had poor generalization on the Perception Test, while TAPIR performed relatively better, and thus it was used as the baseline.

By analyzing TAPIR's motion trajectory predictions, we have identified two issues, as illustrated in Figure \ref{fig: problem} (b): \textbf{i) Static Point Jitter:} In static camera videos, points occasionally showed slight positional jitter despite being static in the Ground Truth Tracker. \textbf{ii) Static Point Pseudo-Following:} While predicting point trajectories, originally static points often shifted due to the coverage of moving objects, either following object movement or drifting in subsequent predictions.

To deal with these challenges, we propose a two-stage approach, which is called TAPIR+. Firstly, a multi-granularity temporal shot motion detection algorithm was designed to distinguish between moving and static camera shots by assessing frame displacement at global and local scales. Secondly, we employed differential processing for videos shot by moving and static cameras. For moving camera videos, we used TAPIR for point trajectory prediction. For static camera videos, we applied a moving object segmentation algorithm to detect motion areas, considering them reliable for point trajectory prediction, while the remaining points were treated as static based on a static baseline.

\vspace{3pt}

%

\vspace{5pt}
\begin{figure*}
        \centering
    \includegraphics[scale=0.50]{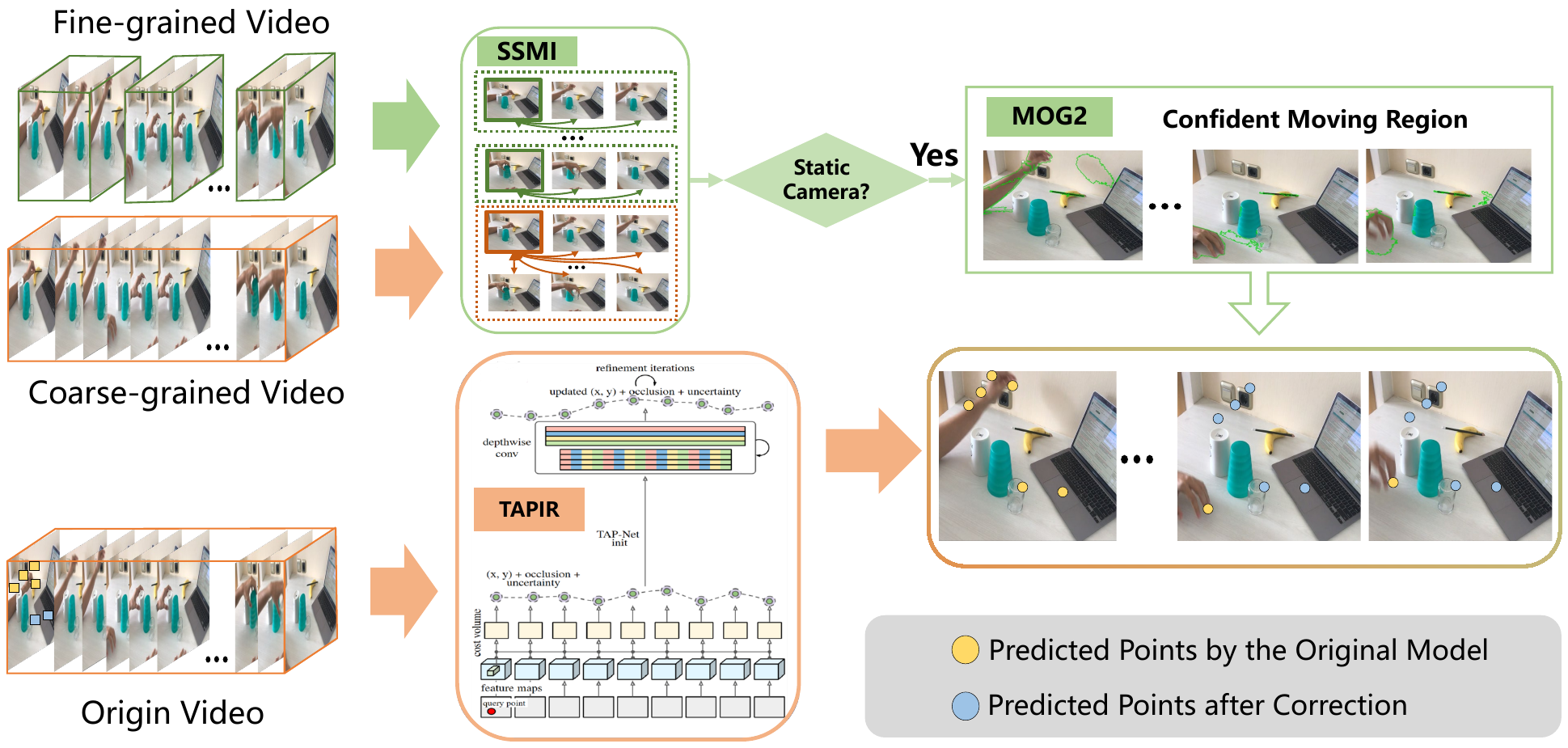}
    \caption{In the framework, the original video is initially fed into the TAPIR model for preliminary point trajectory predictions. Subsequently, the SSMI metric is computed for both the entire video and its fine-grained clips. Analyzing this metric helps distinguish whether the video was shot with a static camera. If the condition is met, a confident moving region is computed for each frame of the video using a motion detection algorithm, thus correcting points that fall outside the region to the initial points. If the condition is not met, the original model's results are directly output.}
    \label{fig: ourframework}
    \vspace{-5pt}
\end{figure*}
\vspace{-5pt}

We introduce a multi-level error mitigation strategy, and its contributions can be summarized as follows:
\begin{itemize}
[itemsep=0pt,parsep=0pt,topsep=0pt,partopsep=0pt,leftmargin=*]
    \setlength{\itemindent}{1.3em}
    \item Multi-granularity temporal shot motion detection algorithm to assess camera shot motion.
    \item Partition-based point trajectory prediction algorithm based on confident moving regions to alleviate jitter in static point predictions and misjudgments caused by occlusion.
\end{itemize}
\section{Method}
\subsection{TAPIR}
In this work, the TAPIR was taken as the base model, which employs a two-stage approach, involving matching and refinement, to independently query and predict fine-grained point trajectories and features based on local information. In contrast to previous TAP methods, TAPIR incorporates uncertainty measurement to determine the reliability of model predictions. After initializing coarse-grained trajectory positions by utilizing the similarity between the features of initial points and the region blocks in each frame. Subsequently, local features are extracted from a neighborhood around the initial estimate and compared to query features at a higher resolution. A temporal depth convolution network is then used for post-processing to refine the similarity, resulting in more accurate position estimates. However, this fine-grained temporal trajectory prediction, which estimates point neighborhoods through sampling, can accumulate errors during the update process. Specifically, TAPIR exhibits jitter in predicting stationary points. Additionally, when query points on a stationary object's surface are occluded by moving objects, their trajectories experience drifting phenomena. 


\subsection{The Proposed TAPIR+}
To solve the aforementioned challenge, we focus on rectifying the tracking of static points in the videos shot by static camera. The proposed method contains two main components as shown in Figure \ref{fig: ourframework}, which will be introduced in the subsequent sections.
\subsubsection{Multi-granularity Camera Motion Detection}
To distinguish between static and moving camera shots in various videos, we devised a multi-granularity temporal shot motion detection algorithm. This algorithm concurrently considers both video-level (coarse-grained) and clip-level segment (fine-grained) analyses, where the original video $\textbf{x}$ is partitioned into multiple 5-second clip segments $\textbf{x}=\{\textbf{c}^1, \textbf{c}^2, ..., \textbf{c}^n\}$, $n$ represents how many clips the video has been divided into. Each frame in videos and clips undergoes a grayscale transformation. In the coarse-grained context, we designate the first frame as the reference frame and compute the \textbf{Structure Similarity Index Measure} (SSIM)~\cite{abs-2006-13846} between subsequent frames and the reference frame. These scores are directly calculated using functions from the \textit{skimage} library. Lower SSIM scores indicate greater dissimilarity between frames. In our experiments, frames with similarity scores falling below a manually set inter-frame similarity threshold ($\lambda_1$) are identified as exhibiting significant differences from the reference frame. We calculate the proportion of frames with substantial dissimilarity to the total number of frames, and if this proportion exceeds a predefined threshold ($\eta$), we preliminarily classify the video as shot by a moving camera: 
\begin{equation}
    \textbf{y}_\textbf{x}^{cg} =\mathbbm{1}\left(\left(\frac{1}{T}\sum_{i}^{T} \mathbbm{1}(SSIM(\textbf{v}_1,\textbf{v}_i)<\lambda_1)\right) > \eta \right)
\end{equation}
where $\textbf{v}_i$ denotes $i$-th frame in video $\textbf{x}$, $T$ is the number of frames, and $\mathbbm{1}$ represents the indicator function. $\textbf{y}_\textbf{x}^{cg}$ reflects the initial results of CMD, where $0$ for static and $1$ for moving. Subsequently, we conduct a fine-grained analysis of these videos. For each clip segment, the initial frame serves as the reference frame, and we similarly calculate SSIM scores between subsequent frames and the reference frame within the clip. The average SSIM score across all frames in the clip is computed as the clip's inter-frame similarity. If any clip, within a video initially categorized as shot by a moving camera, exhibits an inter-frame similarity lower than a designated threshold ($\lambda_2$), we conclusively classify it as shot by a moving camera; otherwise, it is classified as recorded with a static camera:
\begin{equation}
    \begin{split}
    \textbf{y}_{\textbf{c}^j}^{fg} &= \mathbbm{1}\left(\frac{1}{T'}\sum_{i}^{T'} SSIM(c^j_1, c^j_i) < \lambda_2\right) ,\\  
    \textbf{y}_\textbf{x}^{fg} &= \textbf{y}_{\textbf{c}^1}^{fg} \vee \textbf{y}_{\textbf{c}^2}^{fg} \vee \ldots \vee \textbf{y}_{\textbf{c}^n}^{fg}
    \end{split}
\end{equation}
where $T'$ is the number of frames contained within a clip, and $j$ denotes $j$-th clip. The final result of camera motion detection for video $\textbf{x}$ can be represented as follows:
\begin{equation}
    \textbf{y}_\textbf{x} = \textbf{y}_\textbf{x}^{cg} \land \textbf{y}_\textbf{x}^{fg}
\end{equation}


\subsubsection{CMR-based point trajectory prediction}
For videos shot by static cameras, we employ distinct processing approaches with a primary focus on enhancing the accuracy of static point tracking. In the case of moving cameras, point trajectory predictions based on the foundational model TAPIR are directly used as the output results. Conversely, for static cameras, we perform frame-wise segmentation to differentiate between moving and static points, placing higher trust in TAPIR's trajectory predictions for moving points and relying on a static tracking baseline for static points. Specifically, we utilize the MOG2 background modeling-based moving object tracking algorithm to segment the contours of regions occupied by moving objects in each video frame, following the approach outlined in \cite{Zivkovic04, ZivkovicH06}. The foreground regions segmented in each frame are referred to as the \textbf{Confident Moving Region} (CMR), and points located within this region are considered trustworthy, with their trajectories predicted using TAPIR. Points outside this region are deemed static, and their predictions are guided by the static tracking baseline to mitigate errors stemming from static point position change. Here cv2 library's pointPolygonTest function is employed to determine whether a target point falls within the Confident Moving Region. Notably, more advanced moving object detection algorithms can potentially replace MOG2 to further enhance the performance.
\subsection{Discussion}
While TAPIR+ demonstrates significant enhancements in videos captured with static cameras, it grapples with certain limitations when applied to moving camera footage. One notable limitation pertains to its inability to effectively account for relatively stationary objects within the frame when the camera is in motion. Consequently, in scenarios where the camera exhibits moving movement, the accuracy of point trajectory predictions does not exhibit performance improvements. Addressing this issue remains an intriguing and open research challenge. Future work may involve developing novel algorithms and methodologies that can better handle moving camera scenarios.

\section{Experiment}
\textbf{Dataset.} As a zero-shot approach, our method relies solely on TAPIR's pre-trained model and weights, with no additional data used for training. Concerning the pre-trained data, TAPIR adjusted camera angles in the generation script of the public MOVi-E dataset to create a customized training dataset named MOVi-F. Both the validation and test sets use the official data provided.


\textbf{Metric.} The evaluation metric for this Task is the Average Jaccard (AJ), proposed in TAP-Vid~\cite{DoerschGMRSACZY22}.

\textbf{Implementation Detail.} All models are assessed using videos with a maximum resolution of $256 \times 256$, and subsequently, all TAP-Vid metrics are calculated within the same resolution of $256 \times 256$. Our approach is founded on zero-shot learning principles and makes use of the pre-trained model made available on the official TAPIR website\footnote{\label{myfootnote}$https://storage.googleapis.com/dmtapnet/tapir\_checkpoint\\\_panning.npy$}. Regarding hyperparameter settings, we respectively configure the similarity threshold as $\lambda_1=0.5$ and $\lambda_2=0.46$, and establish a frame dissimilarity ratio of $\eta=0.5$.


\begin{table}[htp]
    \centering
    \begin{tabular}{cccccc}
    \toprule
    Method & AJ (static) & AJ \\
    \hline
    Baseline* & -  & 42  \\
    Cotracker & 20.02 & 18.44 \\
    TAPIR &  44.40 & 43.24  \\
    TAPIR+ & 47.19 & 45.78  \\
    \toprule
    \end{tabular}
\caption{Comparison method. *: The experimental data is sourced from official authorities. static: The videos are shot by static cameras. }
\label{tab: compare}

\end{table}

\textbf{Comparison Methods Result.} Table \ref{tab: compare} shows the AJ score performance, from which we can observe that: 1) Cotracker exhibits poor performance in the competition dataset, likely due to its limited model generalization capability. 2) TAPIR+ demonstrates superior performance in static camera shots. TAPIR+ outperforms other TAP methods by achieving about 2.79 performance improvements. This phenomenon exhibits that TAPIR+ offers improved performance in mitigating the jitter and drift of static points.

\textbf{Ablation Study.} To analyze the contribution of each component in TAPIR+, we conduct more ablation studies on the test set of the competition. The Average Jaccard score after adding components is demonstrated in Table \ref{tab: abl}. CMD denotes the Camera Motion Detection. It simply replaces the predicted trajectories in the static camera video with the results of the static baseline. The results for static camera videos reveal the prevalence of numerous static points and are influenced by the error accumulation of the model. Consequently, the model's performance in predicting outcomes in these videos falls short of that achieved by the static baseline. Based on CMD, CMR means a global-scale criterion where target points are considered moving if they appear in the Confident Moving Region, relying on TAPIR for subsequent predictions; otherwise, the points' position is replaced with predictions from the static baseline. The results indicate that each component contributes to improvement, and fine-grained judgment of trajectory points in static camera video can lead to further enhancements.
\begin{table}[htp]
    \centering
    \begin{tabular}{cccccc}
    \toprule
    Method & AJ (static)& AJ \\
    \hline
    TAPIR &  44.40 &   43.24  \\
    + CMD & 45.57 &  44.31 \\
    \hspace{2ex}+ CMR &46.69& 45.33 \\
    \hspace{3ex}+ CMR\textdagger & 47.19 & 45.78  \\
    \toprule
    \end{tabular}
\caption{Ablation experiment. CMR\textdagger: the Temporary CMR, denoting a frame-level criterion where target points in the current frame's CMR follow TAPIR for the current prediction and undergo re-evaluation in subsequent frames.} 
\label{tab: abl}
\end{table}

\textbf{Parameter Sensitivity Analysis.} To verify the sensitivity of parameters, we conduct more experiments by tuning the crucial dissimilar frame percentage parameter $\eta$. Table \ref{tab: para} shows the AJ score on the test set of the competition. TAPIR+ achieves the best AJ score with $\eta$ = 0.5, which indicates that TAPIR+ can effectively distinguish between a moving camera video and a static camera video, handle the jitter and drift of static points, and improve the AJ score in static camera videos.
\begin{table}[htp]
    \centering
    \begin{tabular}{cccccc}
    \toprule
    $\eta$ & AJ (static)& AJ (moving) & AJ \\
    \hline
     0.85 & 47.19 & 27.82 & 45.42 \\
     0.8 &  47.06 &  27.82  & 45.31  \\
     0.5 & 47.19 & 31.72 & 45.78 \\
    \toprule
    \end{tabular}
\caption{Parameter sensitivity analysis. moving: The videos are shot by moving cameras.}
\label{tab: para}
\end{table}

\begin{figure}
    \centering
    \includegraphics[width=1\columnwidth]{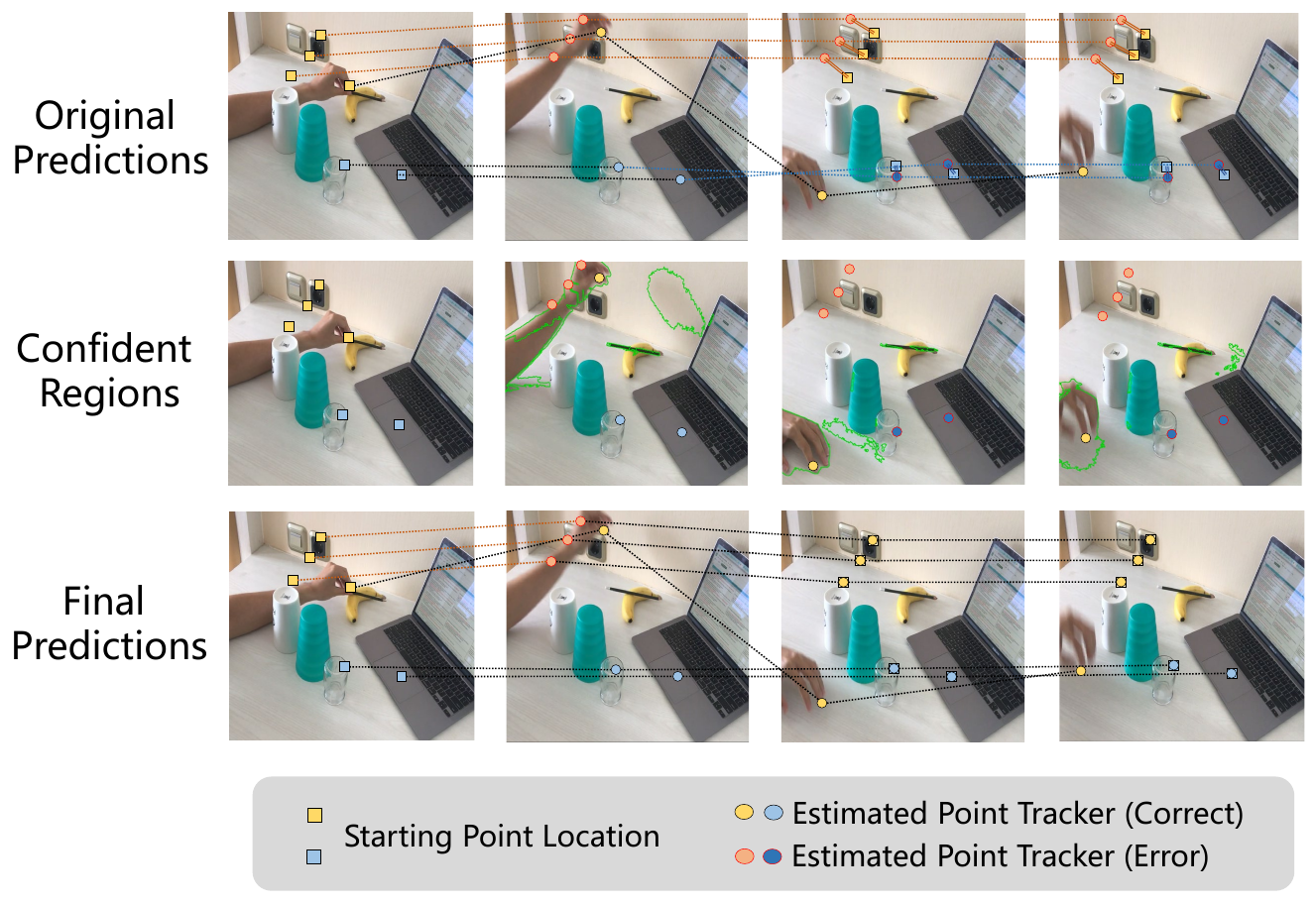}
    \caption{Original Predictions: Predictions from the original model. Confident moving regions: Predictions of confident moving regions using our method. Final Predictions: The final predictions are obtained by combining the original model with our approach.}
    \label{fig: case}
\end{figure}

\textbf{Case Analysis.} 
As shown in Figure \ref{fig: case}, we visualized the trajectories of certain points in the video by applying TAPIR+ to the prediction results, We can observe that in the predictions made by the original model, there are point deviations due to cumulative errors in the model. By identifying confident moving regions in the video, we were able to effectively correct the positions of most drifting points.

\section{Conclusion}

This report summarized our solution for the Point Tracking Task in the ICCV 1st Perception Test Challenge 2023. Our approach was based on camera moving detection and moving object identification to distinguish and stabilize static points in static camera video, thereby effectively enhancing the overall performance of single-point object tracking.

\normalem
{\small
\bibliographystyle{ieee_fullname}
\bibliography{PT_new}
}
\end{document}